\def\year{2022}\relax

\documentclass[letterpaper]{article} 
\usepackage{aaai22}  
\usepackage{times}  
\usepackage{helvet}  
\usepackage{courier}  
\usepackage[hyphens]{url}  
\usepackage{graphicx} 
\usepackage{natbib}  
\usepackage{caption} 
\DeclareCaptionStyle{ruled}{labelfont=normalfont,labelsep=colon,strut=off} 
\usepackage{algorithm}
\usepackage{algorithmic}

\usepackage{multirow}
\usepackage{newfloat}
\usepackage{listings}
\floatstyle{ruled}
\newfloat{listing}{tb}{lst}{}
\floatname{listing}{Listing}
\title{ALEBk: Feasibility Study of Attention Level Estimation via Blink Detection applied to e-Learning}
\author {
    Roberto Daza,
    Daniel DeAlcala,
    Aythami Morales,
    Ruben Tolosana,
    Ruth Cobos,
    Julian Fierrez 
}
\affiliations {
    School of Engineering, Autonomous University of Madrid\\
    \{roberto.daza, daniel.dealcala, aythami.morales, ruben.tolosana, ruth.cobos, julian.fierrez\}@uam.es
    
}

\usepackage{bibentry}

\begin{document}

\maketitle

\begin{abstract}
This work presents a feasibility study of remote attention level estimation based on eye blink frequency. We first propose an eye blink detection system based on Convolutional Neural Networks (CNNs), very competitive with respect to related works. Using this detector, we experimentally evaluate the relationship between the eye blink rate and the attention level of students captured during online sessions. The experimental framework is carried out using a public multimodal database for eye blink detection and attention level estimation called mEBAL, which comprises data from $38$ students and multiples acquisition sensors, in particular, \textit{i)} an electroencephalogram (EEG) band which provides the time signals coming from the student's cognitive information, and \textit{ii)} RGB and NIR cameras to capture the students face gestures. The results achieved suggest an inverse correlation between the eye blink frequency and the attention level. This relation is used in our proposed method called ALEBk for estimating the attention level as the inverse of the eye blink frequency. Our results open a new research line to introduce this technology for attention level estimation on future e-learning platforms, among other applications of this kind of behavioral biometrics based on face analysis.
\end{abstract}

\noindent 

\begin{figure*}[!t]
 \centering
  \includegraphics[width=\linewidth]{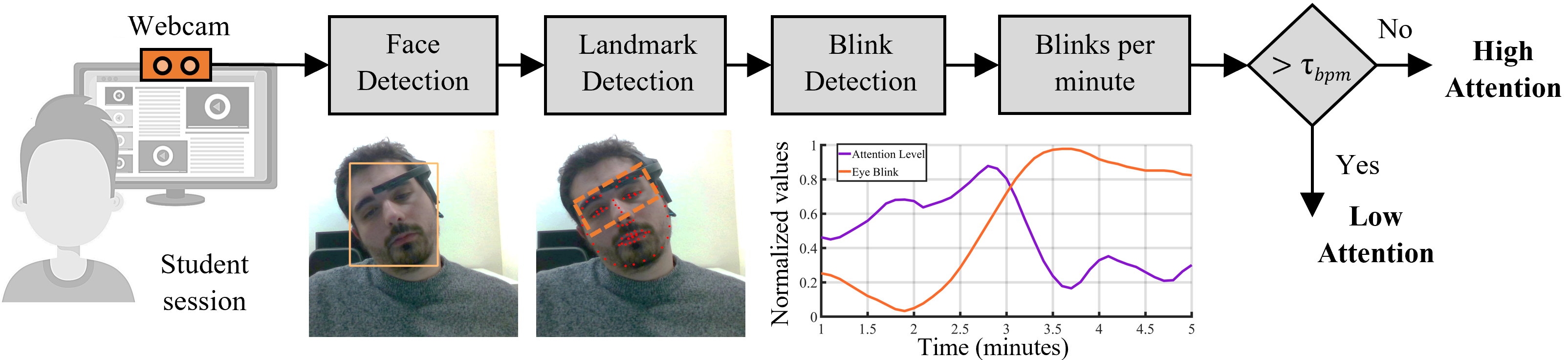} 
  \caption{Architecture of the proposed attention level estimation approach based on blink frequency detection.}
   \label{fig:block}
\end{figure*}

\begin{figure*}[t]
 \centering
  \includegraphics[width=0.7\linewidth]{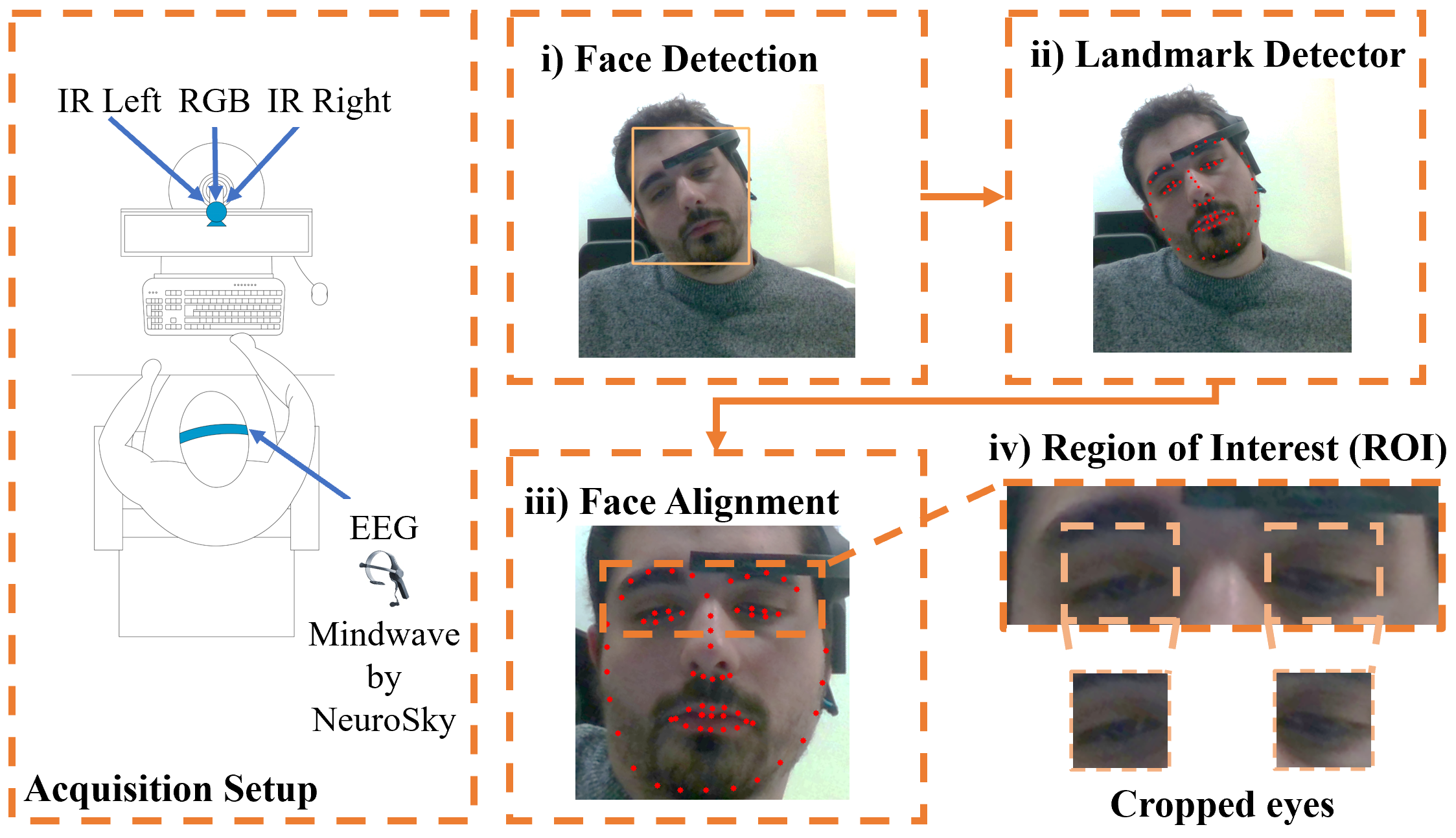} 
  \caption{On the left side it is shown the acquisition setup used on mEBAL. On the right it is shown the pipeline of the  proposed approach for the localization of the region of interest: Face Detection, Landmark Detection, Face Alignment, and ROI Extraction.}
   \label{fig:ROI}
\end{figure*}

\section{INTRODUCTION}

During the last decades, e-learning and virtual education platforms have had a period of high growth \cite{chen2018research}, becoming an essential part in the academic strategy of most important higher education institutions in the world.

Although e-learning presents many advantages \cite{bowers2015students}, it also hands over certain challenges compared to the traditional face-to-face education. One of the most important aspects is the absence of a direct contact between teachers and students. As a result, recent e-learning platforms \cite{hernandez2019edbb} allow to incorporate novel technologies to estimate different factors such as the attention level \cite{daza2020mebal}, the heart rate \cite {javier2020comparative}, the emotional state \cite{shen2009affective}, and the gaze and head pose \cite{asteriadis2009estimation}.


Among these behavioral modalities, the estimation of the student attention level is especially interesting for e-learning platforms \cite{pengpredicting}. This information could be used, for example, to: \textit{i)} adapt dynamically the environment and content \cite{2018_INFFUS_MCSreview1_Fierrez, nappi2018context} based on the attention level of the user, and \textit{ii}) improve the educational materials and resources with a posterior analysis of the e-learning sessions, e.g. detecting the most appropriate types of contents for a specific student and adapting the general information to her \cite{Fierrez-Aguilar2005_AdaptedMultimodal, Fierrez-Aguilar2005_BayesianAdaptation}.

On the other hand, since the $70$s, several studies have analyzed the relationship between the eye blink rate and the cognitive activity of the person such as the attention level \cite{bagley1979effect, holland1972blinking}. These studies suggest that lower eye blink rates can be associated to high attention periods while higher eye blink rates are related to lower attention levels. Consequently, eye blink detection can be a very useful tool to estimate the students attention level and for improving e-learning platforms. However, the potential use of automatic eye blink detectors to infer the attention level of users has not been evaluated in realistic scenarios. The existing studies are limited to initial qualitative analyses that suggest the correlation between signals \cite{daza2020mebal}. Furthermore, eye blink detection can be used for fraud/cheating/lies detection as the eye blink rate decreases when cognitive demand increases \cite{mann2002suspects}.

Eye blink detection methods can be classified into two main groups: \textit{i)} Image-based approaches:  in this case the methods classify each individual frame as \textit{opened}, \textit{closed}, or the degree of eye closure \cite{anas2017online, remeseiro2015automatic}. \textit{ii)}  Video-based:  these approaches consider the temporal information obtained from the entire sequence of frames \cite{soukupova2016eye, hu2019towards}. 

The main contributions of this work are:

\begin{itemize}
    \item To the best of out knowledge, we present the first feasibility study of attention level estimation based on blink frequency detection (see Fig. \ref{fig:block}). 
    \item A novel architecture for blink detection based on Convolutional Neural Networks (CNN). This architecture has been trained using the largest public blink database up to the date. 
     \item Our blink detector has been evaluated over a public eye blink database \cite{hu2019towards}, considering \textit{in-the-wild} scenarios. Our experiments suggest that it is possible to detect high and low attention periods with a moderate accuracy around $74\%$. 
\end{itemize}

The rest of the paper is organized as follows: Section $2$ presents the materials and methods, including the databases and the proposed blink detection approach. Section $3$ shows the experiments and results. Finally, remarks and future work are drawn in Section $4$.

\section{MATERIALS AND METHODS}

\subsection{Database for Attention Level Estimation: mEBAL}
 
 In this work the mEBAL database\footnote{\url{https://github.com/BiDAlab/mEBAL}} is employed \cite{daza2020mebal}. This database was acquired with an experimental e-learning platform for remote education assessment called edBB \cite{hernandez2019edbb}. Students had to perform different tasks including an enrollment form, logic questions, and visual questions for an average total duration of $23$ minutes. mEBAL comprises data acquired from multiple channels including cameras and cognitive signals. Regarding the task of eye blink detection,  mEBAL comprises $6$,$000$ samples ($3$,$000$  blinks and $3$,$000$  no-blinks) from $38$ different students. Each eye blink sample is composed of $19$ frames with $3$ cameras (2 NIR cameras and 1 RGB camera). This database is $8$ times larger than related public eye blink databases. 
 
 In addition, mEBAL includes cognitive signals from an electroencephalogram (EEG) headset by NeuroSky\footnote{\url{https://store.neurosky.com/pages/mindwave}}, which measures the power spectrum density on $5$ channels of electroencephalographic information ($\alpha, \beta, \gamma, \delta, \theta$). More specifically, the EEG data obtained are used to measure the voltage signals produced by synaptic excitations of the pyramidal cells dendrites in the top layer of the brain cortex \cite{li2009towards}. The signals intensity is produced mainly by the number of neurons and fibers that are fired synchronously and it needs thousands/millions to be able to record information \cite{hall2020guyton}. 
 
 


\begin{figure*}[t]
 \centering
  \includegraphics[width=\linewidth]{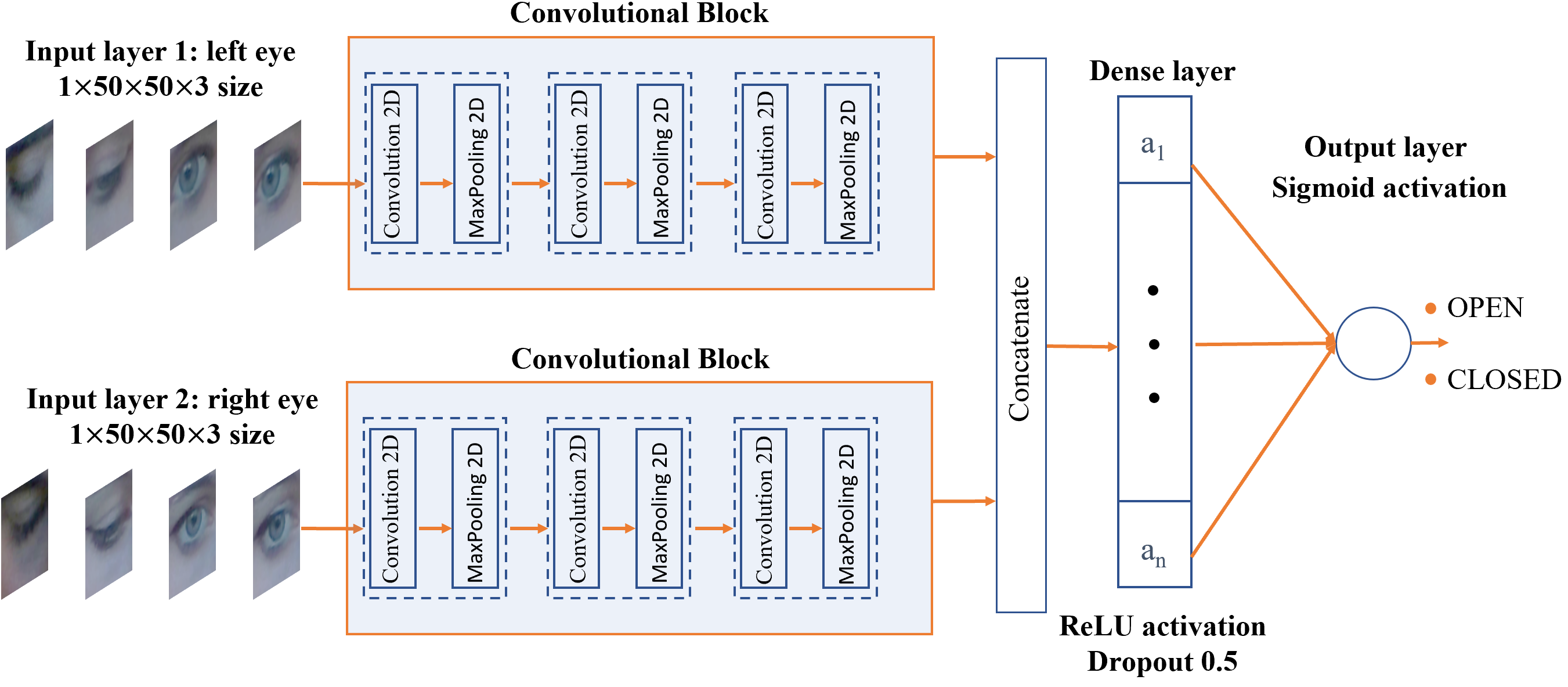} 
  \caption{Architecture of the proposed CNN eye blink detector based on two convolutional blocks (one for each eye).}
   \label{fig:architecture}
\end{figure*}

EEG information is considered one of the most important, effective, and objective measures in estimating the attention level \cite{chen2018effects, li2011real}, since these signals are sensitive to the mental effort and cognitive work, which varies significantly between activities like learning, lying, perception, stress, etc. The resulting EEG information consists of $5$ signals in different frequency ranges. Previous studies in the neurology field have carried out extensive research on EEG signals and their relationship with intellectual and cognitive activities \cite{hall2020guyton, lin2018mental, chen2017assessing, li2009towards}, proving that EEG signals are rich in cognitive information.

Besides, from the official SDK of NeuroSky, the EEG band provides: the attention level, the meditation level, and the temporal sequence with the eye blink strength. These parameters are estimated from the $\delta$ ($<4$Hz), $\theta$ ($4$-$8$ Hz), $\alpha$ ($8$-$13$ Hz), $\beta$ ($13$-$30$ Hz), and $\gamma$ ($>30$ Hz) signals. The attention and meditation attributes have a value ranging from $0$ to $100$.  The sampling rate of the band is $1$ Hz. 

We used the EEG headset to capture the cognitive activity of the student and the ground truth of the eye blinks. We first consider the EEG-estimated eye blinks as candidates for ground truth. Then, a posterior human intervention discards false positives. The remaining eye blinks are considered as a high-quality ground truth.





\subsection{Blink Detection Approach}

We propose an automatic eye blink detector based on Convolutional Neural Networks. It has been trained  combining the visible (RGB) and near infrared (NIR) spectrum. Note that NIR images have only been used during training. In the inference phase the images used are exclusively acquired from the RGB camera (webcam). The eye blink detector is characterized by two main phases: localization of the region of interest and eye blink detection. 

\subsubsection{Region of Interest Localization:} Fig. \ref{fig:ROI} shows our proposed approach for the localization of the region of interest. This approach has four phases: \textit{i}) face detection, \textit{ii}) landmark detection, \textit{iii}) face alignment, and \textit{iv}) eye cropping.

For the face detection we consider the RetinaFace Detector \cite{deng2020retinaface}. Two popular landmark detectors were evaluated based on the iBUG 300-W dataset \cite{sagonas2016300} and Convolutional Pose Machines (CPM) \cite{dong2018supervision}. Our initial experiments over the mEBAL dataset showed a superior performance of the detector based on CPM with improvements around a $2$\% of detection accuracy. The dlib library is used to make the face alignment. The face alignment serves to normalize the face pose. We apply a normalization trying to have eyes on a horizontal line \cite{2015_FSI_FacialSoftBio_Pedro}. The bounding boxes of the eyes are cropped for both eyes using the detected landmarks \cite{2013PTomeFSI_FacialRegions}. The resulting cropped regions are resized to images of $50 \times 50 \times 3$.

\subsubsection{Eye Blink Detection:} Fig. \ref{fig:architecture} shows the architecture proposed for the eye blink detector. The architecture is inspired in the popular VGG16 neural network model \cite{simonyan2014very}, considering two parallel CNNs trained from scratch (one for each eye). The architecture has two input layers of $50 \times 50 \times 3$ size. Each input layer is connected with a convolutional block. All convolutional blocks have the same architecture consisting of $3$ convolutional layers with ReLU activation ($32/32/64$ filters of size $3 \times 3$), and $3$ max pooling layers between them. The output of the first and second convolutional blocks are concatenated. Finally, we consider a dense layer of $64$ units with ReLU activation, and the final output layer with one unit (sigmoid activation). In addition, dropout ($0.5$) is used to reduce overfitting. The batch size is set up to $32$. Adam optimizer is considered with default parameters ($0.001$ learning rate). The network is trained as a binary classifier (open or closed eyes) using  two input images (cropped left eye and cropped right eye) to make the decision. We use mEBAL to train the network with the eye blinks of the RGB and NIRs cameras.

\subsubsection{Attention Level Estimation:} 
\begin{figure}[t]
 \centering
  \includegraphics[width=\linewidth]{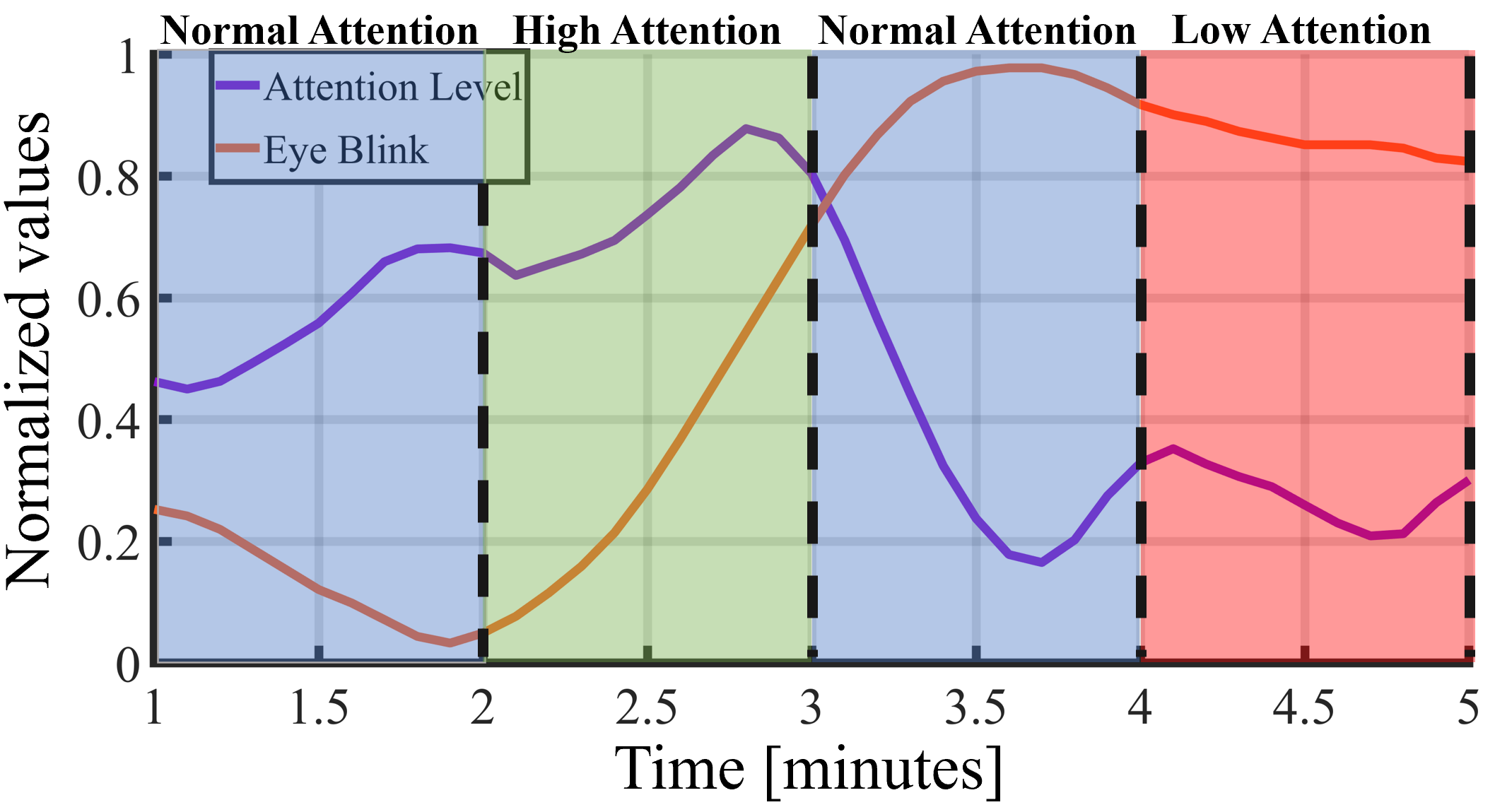} 
  \caption{Normalized Attention Level and Eye Blink rate obtained by the NeuroSky headset. We classify each minute according to three different categories of attention: high, low, and normal attention. The signals show the inverse relationship between attention level and blink rate.}
   \label{fig:Example}
\end{figure}

Our target is to analyze the relation between the eye blink rate and the cognitive activity in a remote education environment. Based on the literature discussed in previous sections, we expect to notice an inverse relationship between the eye blink rate and the level of attention. For the analysis done in this work, and as indicated previously, we use the data included in mEBAL, more specifically the attention levels captured by the EEG band and the eye blinks detected. First, for each user we synchronize the attention and eye blink signals. Then, we calculate the average attention level per minute and the eye blink rate as bpm (blinks per minute). Both, the attention level and the blink rate are obtained by the EEG headset and they will be used as ground-truth for the experiments. Fig. \ref{fig:Example} shows the normalized values of the attention signal and blink rate for a $5$-minutes session. Instead of trying to predict the exact value of the attention or the eye blink rate per second, we argue that it is possible to predict high/low sustained levels of attention over some time using exclusively the images captured by the webcam.

\begin{figure}[t]
 \centering
  \includegraphics[width=\linewidth]{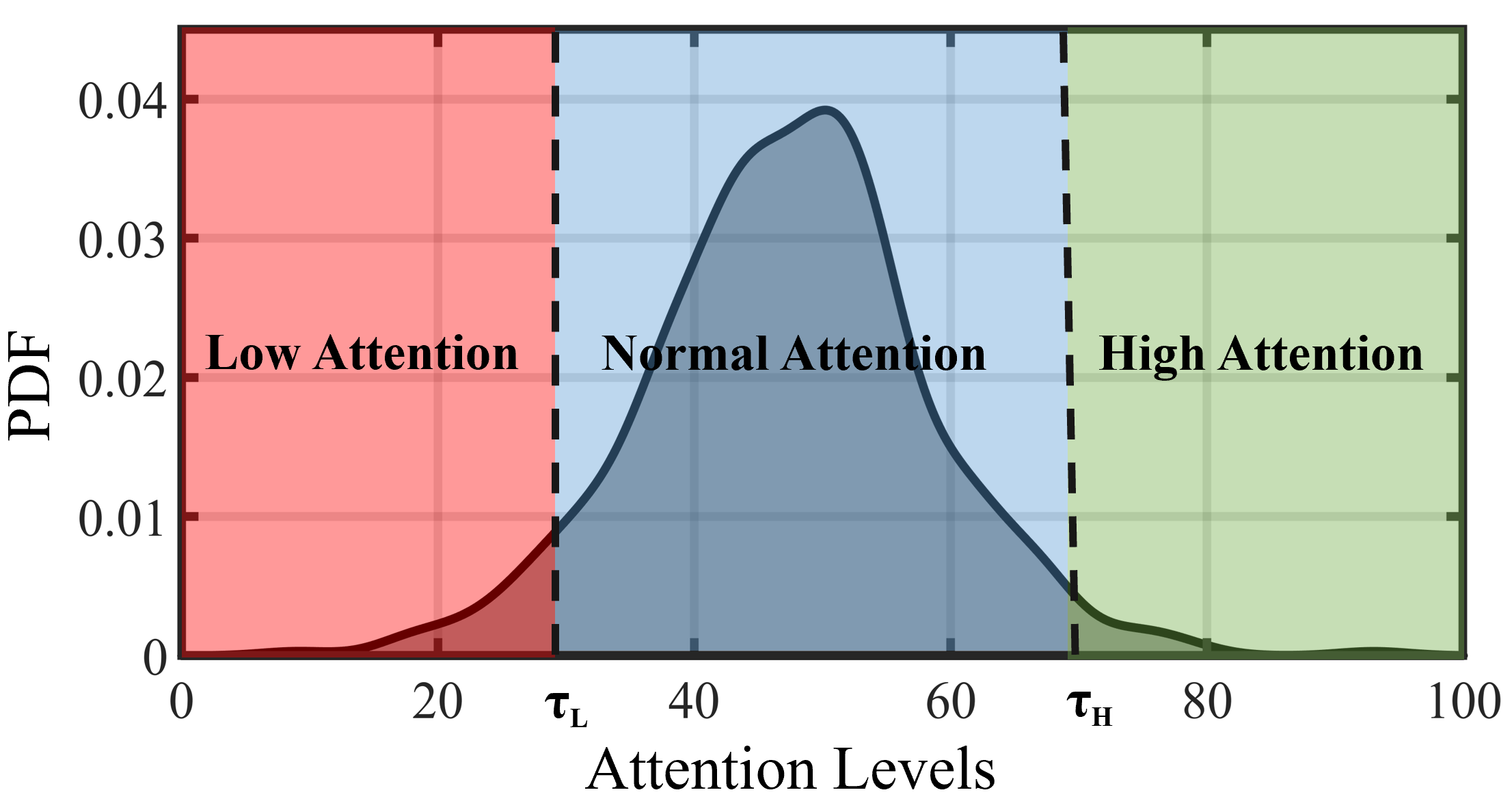} 
  \caption{Probability Density Function of the attention levels provided by the SDK of the NeuroSky band in the mEBAL database.}
   \label{fig:PDF}
\end{figure}

Fig. \ref{fig:PDF} shows the Probability Density Function (PDF) of the attention levels provided by the SDK of the NeuroSky band considering all the students and sessions of the database. Most of the time, the attention level remains in a mean value around $50\%$. The main aim of our experimental framework is to detect periods of high and low attention levels, and therefore, we consider three different groups: high attention (attention higher than a threshold $\tau_H$), low attention (attention lower than a threshold $\tau_L$), and normal attention (attention in between). For the experiments conducted here we define two symmetric thresholds with $\tau_H=100-\tau_L$ and $\tau_L=\mu-\Delta$. Where $\mu$ is the average of the attention level ($\mu=47.27$ in the mEBAL dataset) and $\Delta$ is a variable that controls how far with respect to the mean attention we position the low and high attention thresholds (see Fig. \ref{fig:PDF}.) According to these two thresholds, we have estimated the attention level (High-Normal-Low) of the students for periods of 1 minute (without overlap). In total we have $660$ values of attention level ($660$ minutes) and the corresponding images from the webcam. These images served to estimate the blink rate using the detector proposed in the previous Section. Then, we experimentally establish a blink rate threshold $\tau_{bpm}$ (see next Section). When the value of the estimated blink rate is above $\tau_{bpm}$, the attention level is classified as low. When the value of the estimated blink rate is below $\tau_{bpm}$, the attention level is classified as high. 


\section{EXPERIMENTS AND RESULTS}

\subsection{Eye Blink Detection Results}

\begin{figure*}[t]
 \centering
  \includegraphics[width=\linewidth]{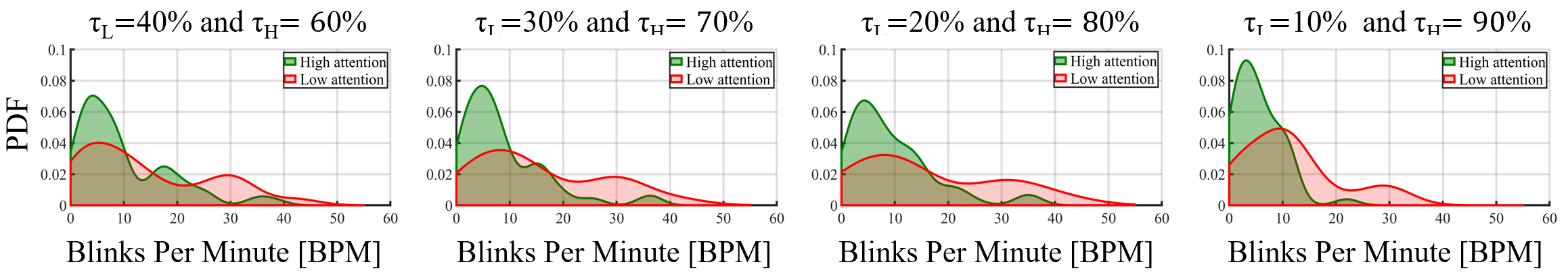} 
  \caption{Probability density function of eye blinks per minute during high and low attention periods (4 different values of $\tau_L$).}
   \label{fig:PDF BLINK RATE}
\end{figure*}

First of all, we evaluate the performance of the proposed eye blink detector. The evaluation is performed with two databases: mEBAL \cite{daza2020mebal} and HUST-LEBW benchmark \cite{hu2019towards}. The mEBAL database is used to train our blink detector and considers a controlled environment, while the HUST-LEBW dataset is obtained in an unconstrained environment. The evaluation on HUST-LEBW allows to evaluate the generalization ability of the proposed eye blink detector to unseen scenarios \cite{2018_TIFS_SoftWildAnno_Sosa, 2018_IntelSys_Trends_Proenca}.

The evaluation in mEBAL is performed with a leave-one-out cross validation protocol. We leave out one user for testing and we train using the remaining users. We repeated the process for all users in the database. The decision threshold was fixed to the point in which the False Positive and False Negative rates in blink detection are equal. 
Our eye blink detector gets a $96.16$\% of accuracy for a total $1$,$094$,$400$  images ($9$,$600\times19\times2\times3$).

HUST-LEBW is very different from mEBAL. It is an unbalanced database that includes only $381$ blink and $292$ no-blink samples. Each sample comprises $13$ frames. We process each image with the eye blink detector proposed. We get a score for each pair of eyes. We have $13$ scores obtained for each sequence of frames (one per frame), the minimum is selected to represent the sample score (minimum scores are more important in the sequence because they are closed  eyes candidates). As in the mEBAL evaluation, we fixed the threshold  to the point in which the False Positive and False Negative rates in blink detection are equal. Table \ref{tab:Comparison} shows the results obtained by our approach and the comparison with previous approaches evaluated over the same HUST-LEBW dataset \cite{chau2005real, drutarovsky2014eye, hu2019towards, morris2002blink, soukupova2016eye, tabrizi2008open, daza2020mebal}. Our method outperforms state-of-the-art eye blink detection algorithms (in terms of F1 score) in an unconstrained environment, which is very different compared to the traditional e-learning environment. In addition, it is interesting to highlight that the proposed approach also outperforms our initial approach presented in \cite{daza2020mebal}. This is produced as both eyes are used as input, giving more information to the network and allowing to reduce the error due to head orientation, illumination, etc.

\begin{table}[t]
\setlength\tabcolsep{7pt}
\renewcommand{\arraystretch}{1.1}
  \caption{Eye Blink detection results on HUST-LEBW dataset. Methods: A=\cite{morris2002blink}, B=\cite{chau2005real}, C=\cite{tabrizi2008open}, D=\cite{drutarovsky2014eye}, E=\cite{soukupova2016eye}, F=\cite{hu2019towards}, G=\cite{daza2020mebal}.}
  \label{tab:Comparison}
 
  \begin{tabular}{c c c c c}
    \hline
     \multirow{1}{1.5cm}{ \centering \bf{Method}}&  \multirow{1}{1.5cm}{\centering \bf{Eye}} &  \multirow{1}{1cm}{\centering \bf{Recall}} & \multirow{1}{1cm}{\centering \bf{Precision}} & \multirow{1}{1cm}{\centering \bf{F1 }} \\
    \hline \hline
     \multirow{2}{1.3cm}{\centering A} & \multirow{2}{1cm}{ \centering Left\\Right}
     & \multirow{2}{1cm}{\centering 0.0164 \\  0.0159}
     & \multirow{2}{1cm}{\centering 0.6667 \\ \textbf{1.0000}}
     & \multirow{2}{1cm}{\centering 0.0320 \\ 0.0313} 
 
 \\ \\ \hline
 \multirow{2}{1cm}{\centering B} & \multirow{2}{1cm}{ \centering Left\\Right}
     & \multirow{2}{1cm}{\centering  0.0164 \\ 0.0000}
     & \multirow{2}{1cm}{\centering \textbf{1.0000} \\   0.0000}
     & \multirow{2}{1cm}{\centering  0.0323 \\  0.0000} 
 \\ \\ \hline
    \multirow{1}{1cm}{\centering C} & \multirow{1}{1cm}{ \centering 2 eyes}
     & \multirow{1}{1cm}{\centering 0.0714 }
     & \multirow{1}{1cm}{\centering 0.4500 }
     & \multirow{1}{1cm}{\centering   0.1233 } 
 \\   \hline
 \multirow{2}{1cm}{\centering D} & \multirow{2}{1cm}{ \centering Left\\Right}
     & \multirow{2}{1cm}{\centering  0.0574 \\ 0.0317}
     & \multirow{2}{1cm}{\centering 0.4118 \\    0.3077}
     & \multirow{2}{1cm}{\centering  0.1007 \\  0.0576} 
\\ \\ \hline
\multirow{2}{1cm}{\centering E} & \multirow{2}{1cm}{ \centering Left\\Right}
     & \multirow{2}{1cm}{\centering  0.3607 \\ 0.3016}
     & \multirow{2}{1cm}{\centering 0.6471 \\    0.5758}
     & \multirow{2}{1cm}{\centering  0.4632 \\  0.3958} 
\\ \\ \hline
\multirow{2}{1cm}{\centering F} & \multirow{2}{1cm}{ \centering Left\\Right}
     & \multirow{2}{1cm}{\centering  0.5410 \\ 0.4444}
     & \multirow{2}{1cm}{\centering 0.8919 \\   0.7671}
     & \multirow{2}{1cm}{\centering  0.6735 \\  0.5628} 
\\ \\ \hline  
\multirow{2}{1cm}{\centering G} & \multirow{2}{1cm}{ \centering Left\\Right}
     & \multirow{2}{1cm}{\centering  0.9603 \\ 0.7950}
     & \multirow{2}{1cm}{\centering 0.6080 \\   0.7348}
     & \multirow{2}{1cm}{\centering 0.7446 \\  0.7637} 
  
 \\ \\ \hline
    \multirow{1}{1cm}{\centering \textbf{Proposed}} & \multirow{1}{1cm}{ \centering 2 eyes}
     & \multirow{1}{1cm}{\centering \textbf{0.9339} }
     & \multirow{1}{1cm}{\centering  0.7533}
     & \multirow{1}{1cm}{\centering  \textbf{0.8339} }  
 \\ \hline
  \end{tabular}
\end{table}


\begin{table}[t]
\centering
\setlength\tabcolsep{7pt}
\renewcommand{\arraystretch}{1.1}
\caption{Attention level in terms of classification accuracy using the mEBAL database. We present results for $10$ different values of $\tau_L$ and $\tau_H$, considering the maximum accuracy and the accuracy obtained operational point where the False Positive and False Negative rates are equal (accuracy=1-EER). $\tau_{bpm}$ = EER threshold}
\begin{tabular}{c c c c }
\hline
\textbf{Att. thresholds $\tau_L$-$\tau_H$} & \textbf{max acc} & \textbf{acc=1-EER}  & \textbf{$\tau_{bpm}$} \\ 
\hline \hline
50\%-50\% & 0.5611   & 0.5295 & 8.13 \\ \hline
45\%-55\% & 0.5680   & 0.5516 & 7.68 \\ \hline
40\%-60\% & 0.5974 	& 0.5658 & 8.01\\ \hline
35\%-65\% & 0.5843 	& 0.5542 & 7.90\\ \hline
30\%-70\% & 0.6585 	& 0.6585 & 8.46\\ \hline
25\%-75\% & 0.6092 	& 0.6050 & 8.28\\ \hline
20\%-80\% & 0.6368	& 0.6211 & 9.00\\ \hline
15\%-85\% & 0.6479 	& 0.6408 & 7.43\\ \hline
10\%-90\% & \textbf{0.7447} 	& \textbf{0.7021} & \textbf{7.10}\\ \hline
5\%-95\%  & 0.6458 	& 0.5833 & 3.77\\ \hline
\end{tabular}
\label{tab:ACC_ATT}
\end{table}

\subsection{Attention Level Estimation Results}

Fig. \ref{fig:PDF BLINK RATE} shows the probability density distributions of the blink frequency in terms of blinks per minute for high and low attention periods, considering four different thresholds $\tau_L$ (from $40\%$ to $10\%$). The results show that the overlap between distributions is higher for the highest values of  $\tau_L$ ($40$\%). Nevertheless, the separability between distributions improves when $\tau_L$ decreases to $20$\% and $10$\%. As it has been theorized in previous studies, during the periods with high attention, the blink rate often has a lower frequency in comparison to low attention periods. However, during the periods with low attention the blink rate is more homogeneously distributed. These results suggest that high attention periods are easier to recognize than low attention periods. 

\begin{figure}[t]
 \centering
  \includegraphics[width=\linewidth]{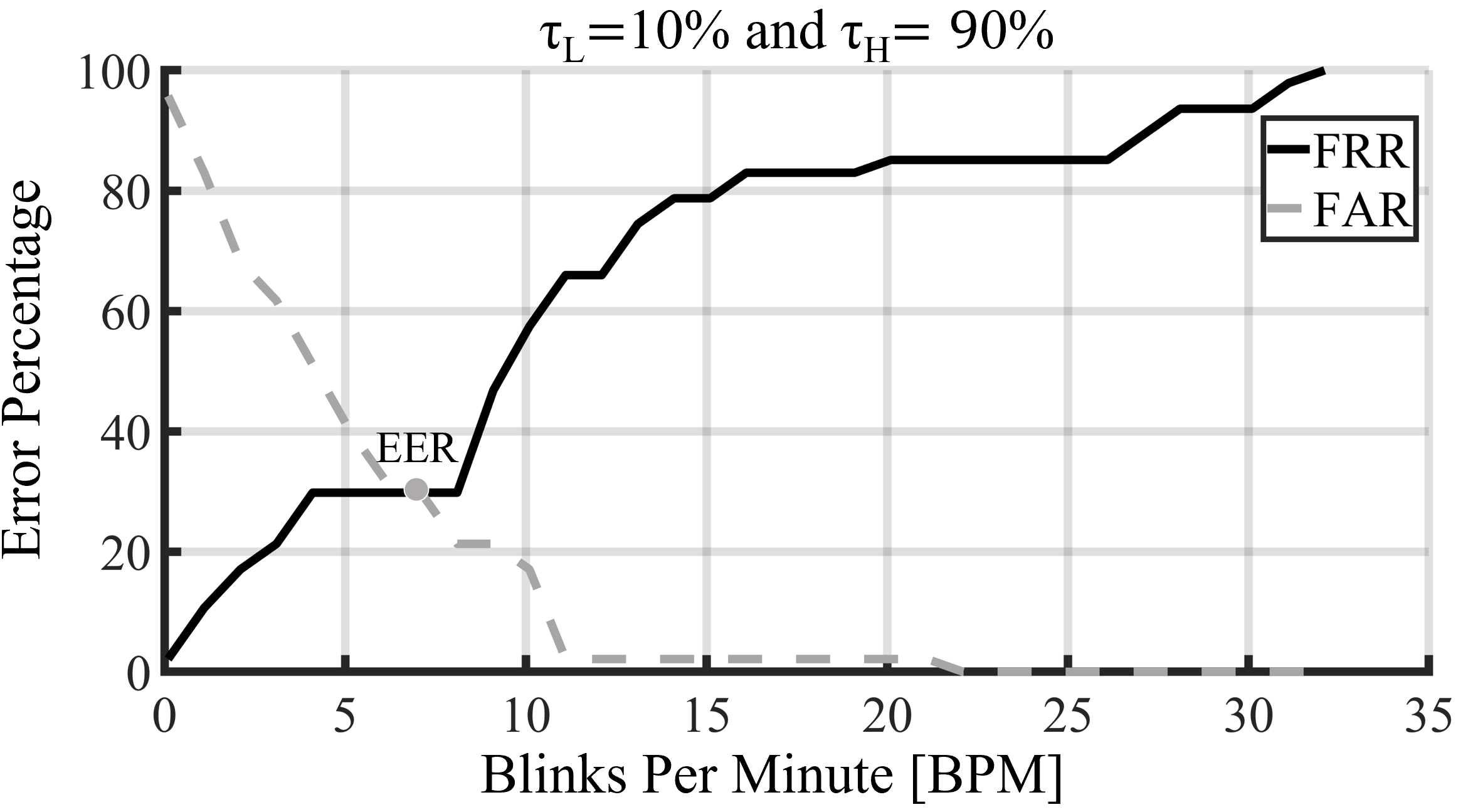} 
  \caption{False Acceptance Rate (FAR) and False Rejection Rate (FRR) versus the blink rate (blinks per minute) for a value of $\tau_L$ equal to $10\%$.}
   \label{fig:FAR AND FRR}
\end{figure}

Table \ref{tab:ACC_ATT} shows the classification accuracy of high and low attention periods for different values of $\tau_L$. The table shows the maximum accuracy and the accuracy for the operational point where the False Positive and False Negative rates are equal. The best results are obtained for the lowest values of $\tau_L$ ($10$\%). In other words, the periods of time with the highest and the lowest attention levels (i.e., extreme cases). For these cases, the blink frequency allows to correctly classify the $70.2$\% of the samples. Fig. \ref{fig:FAR AND FRR} shows the False Acceptance Rate and the False Rejection Rate for this case ($\tau_L=10$\%). According to these results, we can observe a certain level of correlation between the level of attention and the eye blink frequency, which is coherent with the literature.

\section{CONCLUSIONS AND FUTURE WORKS}

We proposed a novel Attention Level Estimation approach based on eye Blink frequency detection, coined ALEBk. Our method is based on a new eye blink detector based on a Convolutional Neural Network (CNN) that has been trained from scratch using RGB and NIR images from the mEBAL database\footnote{\url{https://github.com/BiDAlab/mEBAL}}. We have evaluated our blink detection approach using two very different databases (mEBAL and HUST-LEBW). Our results have outperformed the state of the art, showing that the architecture presented in this paper can be used in real applications like remote education \cite{hernandez2019edbb}. Additionally, based on the studies that correlate the eye blink rate with the cognitive activity, we have analyzed the inverse relationship between the eye blink rate and the attention level in a remote education environment. Our experiments suggest that there is certain correlation between the attention levels and the eye blink rate. We have obtained ca. $74$\% of maximum accuracy in the detection of periods with high and low attention levels.

Future studies should explore other deep learning architectures based on Convolutional Long Short-Term Memory (ConvLSTM) neural networks or other architectures combining both short- and long-term information \cite{2021_AAAI_DeepWriteSYN_Tolosana} to take advantage of the time dimension to better detect the blinking. Additionally, multimodal approaches \cite{fierrez06phd} based on other students behaviors apart from the eye blink frequency should be explored to improve the classification accuracy \cite{pengpredicting}, including the rest of the inputs commonly considered in a remote e-learning session: keystroking \cite{2016_IEEEAccess_KBOC_Aythami, morales2016kboc}, mouse \cite{2022_PR_BeCAPTCHA-Mouse_Acien}, touchscreen interaction \cite{2018_TIFS_Swipe_Fierrez}, or other forms of human-computer interaction signals \cite{2020_CDS_HCIsmart_Acien}.

\section{Acknowledgments}

Support by projects: TRESPASS-ETN (MSCA-ITN-2019-860813), PRIMA (MSCA-ITN-2019-860315), BIBECA (RTI2018-101248-B-I00 MINECO/FEDER), BBforTAI (PID2021-127641OB-I00 MICINN/FEDER), and by edBB (UAM). R. Daza is supported by a FPI fellowship from the Spanish MINECO/FEDER.

\bibliography{aaai22}

\end{document}